\documentclass{article}

\PassOptionsToPackage{numbers,compress}{natbib}

\usepackage[dblblindworkshop,preprint]{neurips_2026}
\workshoptitle{AXIOM 2026: Foundations of Efficient Deep Learning}

\makeatletter
\renewcommand{\@noticestring}{%
  Accepted at AXIOM 2026: Foundations of Efficient Deep Learning
  (NeurIPS 2026 Workshop).%
}
\makeatother

\usepackage[utf8]{inputenc}
\usepackage[T1]{fontenc}
\usepackage{booktabs}
\usepackage{amsmath,amssymb,amsfonts}
\usepackage{nicefrac}
\usepackage{microtype}
\usepackage{xcolor}
\usepackage{graphicx}
\usepackage{subfig}
\usepackage{algorithm}
\usepackage{algpseudocode}
\usepackage{hyperref}
\usepackage{url}

\title{CAGE-NAS: Certified Functional Descent for Efficient Model Growth}

\author{%
  Santiago Florido Gomez \\
  Institut Polytechnique de Paris \\
  Paris, France \\
  \texttt{santiago.florido@ensta-paris.fr}
  \And
  St\'ephane Rivaud \\
  Inria Saclay, TAU Team \\
  Paris, France \\
  \texttt{stephane.a.rivaud@inria.fr}
}

\begin{document}

\maketitle

\begin{abstract}
The progressive growth of neural networks requires deciding when the current representation remains sufficient for optimization and when it should be expanded. \textbf{CAGE-NAS} formulates this decision in function space through an admissibility criterion on approximations of the functional gradient. As long as a representation enables a certified Functional Gradient Descent step, the architecture remains fixed; when the criterion fails, a function-preserving expansion is applied and the resulting representation is evaluated again. As the main instance, we study the family induced by the tangent space,
using a regularized projection of the functional gradient. In a controlled setting with exact certification, CAGE-NAS produces architectures positioned above the $99.8$th performance percentile by held-out RMSE among all admissible alternatives within the same parameter budget, without enumerating them during the growth trajectory.

\end{abstract}

\section{Introduction}
\label{sec:introduction}

Neural scaling laws \cite{bahri2021explaining} show that the performance of neural models improves predictably as model capacity, data, and available compute increase, a dynamic that has favored the use of increasingly larger architectures to improve performance on common benchmarks \cite{desislavov2023trends}. 
In this context, Neural Architecture Search (NAS) \cite{elsken2019neural} addresses the automated selection of architectures to allocate model capacity more efficiently. Progressive growth methods incorporate growth as a constructive strategy for adding expressive capacity to the model, which raises a key design problem: determining when the current representation is no longer adequate for optimization and should therefore be modified. The solutions proposed to this problem reveal a more general need: training and growth should be governed by a common condition on the ability of the current representation to continue following the optimization direction.

In this work, we address this decision from the perspective of optimization in function space, where the neural representation induces adaptive families of functional-gradient approximations. The current representation is considered adequate as long as at least one of these families produces a realizable direction sufficiently close to the functional gradient to preserve the descent and convergence conditions of functional gradient descent \cite{csillag2026functional}. Based on this principle, we formulate CAGE-NAS as a constructive scheme that connects training and growth through a common condition of representational sufficiency.
\section{Previous Work}
\label{sec:Previous work}

Constructive NAS methods have proposed different mechanisms to progressively adapt the architecture , 
GradMax \cite{evci2022gradmax} uses gradient information to initialize newly added neurons, 
NORTH \cite{maile2022north} introduces specific criteria for \emph{when, where,} and \emph{how to grow}, Firefly \cite{wu2020firefly} formulates a local search over architectural modifications, and FRAGrow \cite{wu2024fragrow} adapts growth timing according to fitting risk. 
However, these criteria do not directly characterize the functional capacity of the current representation to follow the optimization direction, a question addressed by TINY \cite{verbockhaven2024tiny} through the notion of an \emph{expressivity bottleneck}. In the extension of TINY \cite{verbockhaven2024tiny} to DAG architectures \cite{douka2025dag}, intermediate training between growth events is shown to push the parameters toward overfitting, while subsequent growth can help escape potential local minima, highlighting the importance of the interaction between the \emph{train} and \emph{grow} dynamics.

Within the framework of an algorithm that enables the unification of optimization directions, the formulation of \emph{Functional Gradient Descent with Adaptive Representations} \cite{csillag2026functional}, establishes convergence conditions when the functional gradient is replaced by adaptive finite-dimensional approximations. Under this framework, the foundations are established for formulating a training and structural growth framework governed by a functional approximation criterion.

\section{CAGE-NAS}
\label{sec:cage-nas}

Let $\mathcal{A}$ be the set of admissible finite architectures. Each $A\in\mathcal{A}$ is associated with an open parameter domain $\Theta_A\subseteq\mathbb{R}^{p_A}$, where $p_A$ is the number of trainable parameters. For a compact input domain, let $\mathcal{H}_n$ denote the empirical Hilbert space associated with the $n$ evaluation samples. The finite-data representation is defined as

\begin{equation}
    \Phi_A : \Theta_A \longrightarrow \mathcal{H}_n,
    \qquad
    \Phi_A(\theta)=\bigl(f_{A,\theta}(x_i)\bigr)_{i=1}^{n}.
    \label{eq:representation_map}
\end{equation}
The global parameter-and-architecture domain is naturally represented by the stratified topological sum
\begin{equation}
    \mathcal{P}=\bigsqcup_{A\in\mathcal{A}}\{A\}\times\Theta_A.
    \label{eq:stratified_parameter_space}
\end{equation}
Let the loss be a functional $\mathcal{L}:\mathcal{H}_n\longrightarrow\mathbb{R}$, assumed to be differentiable at $F\in\mathcal{H}_n$. Its functional gradient is then $r=\nabla_F\mathcal{L}(F)$. Following \cite{csillag2026functional}, the ideal update $-r$ of a functional gradient descent step cannot, in general, be implemented directly through the current finite-dimensional representation. Therefore, the objective is to find an approximation $g_t\in\mathcal{B}$ of the functional gradient, induced by $\Phi_A$, that enables gradient descent steps to be executed through adaptive representations, where $\mathcal{B}$ denotes the Banach space containing $\mathcal{H}_n$. In the finite-data setting considered here, the Banach and Hilbert norms are equivalent on $\mathcal{H}_n$. The approximation must necessarily satisfy the relative-error criterion. Moreover, the update steps must be performed in such a way that the loss function converges to a stationary point that is also a global minimizer. Since the structural modification replaces the architecture and, consequently, modifies the parameter space by replacing it with one of a different dimension, the growth operation has to be function-preserving. Therefore, it only increases the set of available functional directions without directly modifying the predictions of the network. Let the convex training loss be the unnormalized squared-error loss~(\ref{app:fgd-mse-properties}),
\begin{equation}
\mathcal{L}(F)=\|F-Y\|_2^2,
\qquad
r(F):=2(F-Y).
\end{equation}
Using the equivalence between the Hilbert and Banach norms on $\mathcal{H}_n$, the relative-error certificate for a representation $g_t$ is defined as
\begin{equation}
\epsilon_t
=
\frac{\|g_t-r_t\|_{B}}
     {\|g_t\|_{B}}
<
\frac{1}{2}.
\label{eq:fgd-relative-error-code}
\end{equation}
For every admissible representation, the sufficient-descent lemma~(\ref{app:fgd-lemma35-proof}) gives
\begin{equation}
\mathcal{L}(F_{t+1})
\leq
\mathcal{L}(F_t)
-
\eta_t q(\eta_t,\epsilon_t)
\|r_t\|_{B}^{2},
\label{eq:fgd-code-descent}
\end{equation}
where $q(\eta_t,\epsilon_t)=1-\eta_t-(1+3\eta_t)\frac{\epsilon_t}{1-\epsilon_t}$, which must satisfy $q(\eta_t,\epsilon_t)>0$. Consequently, the learning rate for each FGD step must satisfy
\begin{equation}
0<\eta_t<
\frac{1-2\epsilon_t}{1+2\epsilon_t}.
\label{eq:fgd-rate-ceiling}
\end{equation}
Furthermore, since the training loss satisfies the PL condition with $\mu=2$, any sequence of committed certified steps satisfies the trajectory-level contraction~(\ref{app:fgd-global-contraction}).

\paragraph{When to grow.}
Given these conditions for Adaptive Representations FGD, in the train-and-grow setting, a set of functional-gradient representations is defined for each network architecture $A$. If none of the functional-gradient representations induced over $\Theta_A$ can certify a step, then the corresponding architecture is unable, at the current functional state, to provide an update direction satisfying the certification criterion, revealing an architectural expressivity bottleneck. Consequently, growth becomes necessary to expand the set of possible directions and increase the expressiveness of the structure.

\paragraph{Where and how to grow.}
In CAGE-NAS, growth is constrained to be function-preserving, making the decision of \emph{where to grow} fundamental for the algorithm~(\ref{app:where-to-grow}). Two methods are considered. The first selects the position $\ell^\star$ in $A$ whose one-neuron expansion $A\oplus e_\ell$, initialized at $\theta_\ell^+$ to preserve the network function, yields the smallest relative error,
\begin{equation}
\ell^\star \in
\arg\min_{\ell}
\epsilon_{\min}
\left(A\oplus e_\ell,\theta_\ell^{+}\right).
\end{equation}

The second follows the expressivity-bottleneck criterion introduced in TINY \cite{verbockhaven2024tiny} and selects the position in the network with the largest expressivity bottleneck. Importantly, the growth operation does not constitute an optimization step itself: the admissible structural transformations are constrained to preserve the current network function, such that $\Phi_{A^{+}}(\theta^{+})=\Phi_A(\theta)$, and therefore the loss remains unchanged at the architectural transition. Once the selected expansion is applied, the certification condition is recomputed on the grown architecture, generating a constructive sequence $A_0\preceq A_1\preceq\cdots$ in which optimization resumes as soon as the enlarged representation can again certify a functional descent step.

\subsection{The damped tangent space approximation as a CAGE-NAS Representation}

Following the considerations in \cite{rudkiewicz2024fogro}, it becomes evident that one of the natural structure-dependent adaptive representations useful in CAGE-NAS arises from the tangent space
\begin{equation}
T_{A,\theta}
=
\left\{
J_{A,\theta}\delta\theta
:
\delta\theta\in\mathbb{R}^{p_A}
\right\},
\label{eq:demeter-fgd-tangent-space}
\end{equation}
where $J_{A,\theta}$ denotes the Jacobian of the finite-data representation, so that $T_{A,\theta}$ characterizes the functional directions generated by infinitesimal parameter variations $\delta\theta$. Therefore, we consider the projection of the functional gradient onto the tangent space, regularized by a damping factor $\lambda\geq 0$. More precisely, it solves the Tikhonov problem
\begin{equation}
u_\lambda
=
\arg\min_{u\in\mathbb{R}^{p_A}}
\left\{
\|J_{A,\theta}u-r\|_2^2
+
\lambda\|u\|_2^2
\right\},
\qquad
g_\lambda
=
J_{A,\theta}u_\lambda.
\end{equation}

In this way, the resulting representation is the functional direction $g_\lambda$, where the damping parameter controls the trade-off between approximation fidelity and the magnitude of the parameter displacement required to realize the direction~(\ref{app:damped-tangent-family}). As the damping parameter approaches the vanishing limit, $g_\lambda$ approaches the orthogonal projection of $r$ onto the tangent space $T_{A,\theta}$, while the residual corresponds to the definition of the \emph{expressivity bottleneck} provided by TINY \cite{verbockhaven2024tiny}. CAGE-NAS evaluates this approximation with the relative-error certificate: an admissible $g_\lambda$ certifies descent under the current architecture, whereas failure of the tangent representation motivates other realizable variations not captured by the first-order approximation of parametric displacements, such as those from the nonlinear parametric displacement family~(\ref{app:nonlinear-parametric-family}), or architectural adaptation.

\section{Empirical Evaluation of CAGE-NAS}
\label{sec:experimental-evidence}

We evaluate certificate-driven growth on a controlled nonlinear regression benchmark~(\ref{app:synthetic-n1024}), using the complete training set for certification. Starting from $[2,2,2]$, CAGE-NAS follows the train-and-grow algorithm~(\ref{app:train-and-grow-algorithms}) under a fixed parameter budget. All admissible three-hidden-layer MLPs within the corresponding parameter budget are screened post hoc~(\ref{app:smaller-architecture-grid}), assessing certified optimization and the discovered architecture's position among feasible alternatives~(\ref{app:controlled-results}). Training and validation contain $1024$ examples each and test contains $8192$; four model seeds expose trajectory sensitivity.

{\setlength{\intextsep}{4pt}
\begin{figure}[H]
  \centering
  \begin{minipage}[c]{0.47\linewidth}
  \centering
  \footnotesize
  \setlength{\tabcolsep}{1pt}
  \begin{tabular}{@{}lrrrr@{}}
    \toprule
    Seed & Architecture & Params. & Test RMSE & Percentile \\
    \midrule
    0 & $[16,19,9]$  & 593 & 0.0642 & 99.95 \\
    1 & $[9,18,18]$  & 586 & 0.0680 & 99.88 \\
    2 & $[13,16,15]$ & 560 & 0.0585 & 100.00 \\
    3 & $[11,17,17]$ & 583 & 0.0544 & 100.00 \\
    \bottomrule
  \end{tabular}
  \end{minipage}
  \begin{minipage}[c]{0.52\linewidth}
  \centering
  \includegraphics[width=\linewidth]{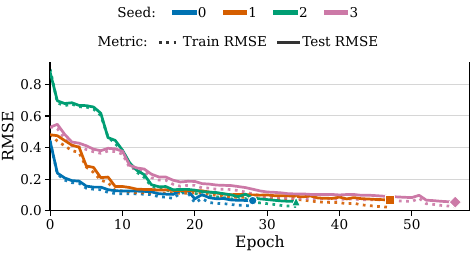}
  \end{minipage}
  \caption{CAGE-NAS discovers compact models in one train-and-grow
  trajectory. The percentile is measured by held-out RMSE against all screened
  models satisfying the strict \(600\)-parameter budget.}
  \label{fig:controlled-results}
\end{figure}}

Figure~\ref{fig:controlled-results} reports mean test RMSE
$0.0613\pm0.0060$. Retraining each discovered architecture from scratch with
the same seed and validation-selected hyperparameters obtains
$0.0700\pm0.0053$ under the fixed-architecture grid~(\ref{app:fixed-architecture-grid}); growth obtains lower RMSE for three of four seeds. This suggests that implementing a compact train-and-grow strategy in terms of the optimization objectives is favorable, identifying at the end of the trajectory not only a useful architecture but also a favorable state within the parameter space for that architecture, without systematic improvements resulting from conventional retraining.
The budget-constrained comparison screens $76{,}646$ architectures and tunes
$12{,}240$ architecture--hyperparameter pairs. CAGE-NAS lies above the
$99.8$th performance percentile by test RMSE for every seed without enumerating those alternatives.
 This demonstrates the high efficiency of the proposed constructive search mechanism, producing architectures that, under the defined budget, are close to the best observed RMSE--capacity trade-offs under the evaluated protocol.

\section{Efficiency boundary and limitations}
\label{sec:efficiency-limitations}

The controlled synthetic benchmark provides focused evidence across four seeds;
classification and real-world tasks are needed to assess broader applicability.
Tangent-family certification has a further boundary. With $s=nc$ evaluated
scalar outputs, $p_A<s$ prevents the empirical Jacobian from spanning the
finite-data function space solely through overparameterization. If $p_A\geq s$
and $J_{A,\theta}$ has full row rank, the tangent family can reproduce any
empirical functional gradient, so a small certificate may reflect interpolation
rather than representational sufficiency. Finally, the reported efficiency
concerns architecture discovery: CAGE-NAS avoids enumerating the
budget-constrained comparison set. Practical scaling depends on the
certification backend; an explicit Jacobian requires $\Theta(sp_A)$ storage and
$\mathcal{O}(sp_A\min\{s,p_A\})$ dense factorization, and certification may
evaluate several expansions, damping values, and solves. Profiling wall-clock,
FLOPs, peak memory, and energy across problem scales remains future work. The
evaluated space is restricted to fixed-depth, three-hidden-layer MLPs and
monotonic neuron additions.

\section{Conclusion}
\label{sec:conclusion}

CAGE-NAS presents a neural architecture growth method that traverses the architectural space through growth as a consequence of the insufficiency of the current representation to follow the desired optimization direction in function space. Under this principle, we propose an adaptation of Functional Gradient Descent with Adaptive Representations, where the adaptive representations arise directly from the capacity of the parameter space of a given architecture to produce a contribution in a specific direction. The tangent family provides a first instantiation of this mechanism, directly relating the accessible functional directions to the local geometry induced by the Jacobian. We show that, in a controlled setting where the Jacobian can be computed exactly and the full dataset is used for certification, the architectures constructed by CAGE-NAS are consistently located in the extreme lower-RMSE tail of the admissible alternatives within the corresponding parameter budget, without requiring their enumeration during the search trajectory. These results provide initial evidence that the adequacy of functional-gradient representations can be used as a criterion to coordinate training and growth within a constructive NAS procedure, leaving as future work its evaluation with other representation families and on larger-scale architectures and problems.

\noindent\textbf{Code availability:}
\href{https://github.com/Santiago23florido/functional-gradient-growth}{Official repository}.

\label{page:last-content}
\ifnum\value{page}>4
  \PackageError{axiom-paper}{AXIOM content exceeds four pages}
    {Shorten the main paper or move optional detail to the appendix.}
\fi

\bibliographystyle{plainnat}
\bibliography{references}

\clearpage
\appendix
\section{Where to grow}
\label{app:where-to-grow}

To determine where to grow as part of the algorithm, two complementary methods
are proposed. Each method is used according to the stage at which the growth
requirement is invoked. The first path identifies which candidate location \(\ell\) within the network
induces the smallest relative error for the current architecture when one
neuron is added at that position. This path is activated during the outer steps
performed by the \texttt{grow\_until\_certify} algorithm.
For each candidate location \(\ell\), write
\(A_\ell^+:=A\oplus e_\ell\), \(F:=\Phi_A(\theta)\), and
\(F_\ell^+:=\Phi_{A_\ell^+}(\theta_\ell^+)\).
The growth operation is function-preserving, so that
the output and residual identities below hold exactly.
\begin{equation}
    F_{\ell}^{+}
    =
    F.
    \label{eq:fgd-function-preserving-candidate}
\end{equation}
Consequently,
\begin{equation}
    r_{\ell}^{+}
    =
    \nabla_F\mathcal{L}(F_{\ell}^{+})
    =
    \nabla_F\mathcal{L}(F)
    =
    r.
    \label{eq:fgd-function-preserving-residual}
\end{equation}
Therefore, all candidates are compared using the same predictions, loss, and
functional gradient. The only quantity that changes is
\begin{equation}
    J_{\ell}^{+}
    =
    D\Phi_{A_\ell^+}(\theta_\ell^+).
    \label{eq:fgd-candidate-jacobian}
\end{equation}
Conceptually, this Jacobian can be written as
\begin{equation}
    J_{\ell}^{+}
    =
    \begin{bmatrix}
        J_{A,\theta} & B_{\ell}
    \end{bmatrix},
    \label{eq:fgd-augmented-candidate-jacobian}
\end{equation}
where \(B_{\ell}\) contains the new Jacobian columns generated by the
parameters of the neuron added at location \(\ell\).
Although initializing the added neuron with values that favor a reduction of
the relative error for the specific architecture may be an interesting
strategy, the growth step would no longer be a simple movement from one
parameter domain \(\Theta_A\) to another parameter domain \(\Theta_{A'}\). Instead, it
would become an additional stage within FGD whose contribution would be
difficult to certify.
For this reason, function-preserving growth is selected as an expansion of the
parameter space between two architectures without directly modifying either
the loss or the network outputs.
Thus, the new tangent space is
\begin{equation}
    T_{\ell}^{+}
    =
    \operatorname{range}(J_{\ell}^{+})
    =
    \operatorname{range}
    \begin{bmatrix}
        J_{A,\theta} & B_{\ell}
    \end{bmatrix}.
    \label{eq:fgd-candidate-tangent-space}
\end{equation}
The corresponding certificate is then computed as
\begin{equation}
    \epsilon_{\lambda,\ell}^{+}
    =
    \frac{
        \|g_{\lambda,\ell}^{+}-r\|_B
    }{
        \|g_{\lambda,\ell}^{+}\|_B
    }.
    \label{eq:fgd-candidate-relative-error}
\end{equation}
Let \(\Lambda\) denote the evaluated damping values. The best certificate for
candidate \(\ell\) is
\[
    \epsilon_{\min}(A\oplus e_\ell,\theta_\ell^+)
    :=
    \min_{\lambda\in\Lambda}\epsilon_{\lambda,\ell}^+.
\]
The selected location is
\begin{equation}
    \ell^\star
    \in
    \arg\min_{\ell}
    \epsilon_{\min}
    \left(
        A\oplus e_{\ell},
        \theta_{\ell}^{+}
    \right).
    \label{eq:fgd-where}
\end{equation}
Thus, ``where'' is determined through a counterfactual comparison of
certificates, rather than through round-robin allocation or the current
layerwise gradient norm.
For compactness, write
\(\epsilon_\ell^+:=\epsilon_{\min}(A\oplus e_\ell,\theta_\ell^+)\).
For \(L\) growable locations, this produces the comparison
\begin{equation}
    \begin{array}{c|cccc}
        \text{candidate}
        &
        A\oplus e_1
        &
        A\oplus e_2
        &
        \cdots
        &
        A\oplus e_L
        \\
        \hline
        \text{score}
        &
        \epsilon_1^{+}
        &
        \epsilon_2^{+}
        &
        \cdots
        &
        \epsilon_L^{+}
    \end{array}
    \label{eq:fgd-candidate-location-comparison}
\end{equation}
The comparison is performed on cloned networks that all start from the same
output \(F\) and use the same data. Therefore, the argmin isolates the effect
of the structural location.
After one candidate is committed, all scores are discarded and recomputed
because the Jacobian, its spectrum, and the set of possible subsequent
additions have changed.
The active location rule compares the relative errors directly.
In contrast, when the outer steps have been completed and the algorithm
requests a growth operation, a different path is followed for the selection of
\emph{where} to grow. Following the manner in which the restricted method of
\citet{douka2025dag} selects a node, the layer with the largest expressivity
bottleneck is identified according to the maximum expressivity-bottleneck
criterion. A neuron is then added to this layer through a
function-preserving construction. A method different from certificate-based selection is used because this is an
outer step whose purpose is to provide an additional degree of freedom for
finding directions with greater expressivity during subsequent executions of
the \texttt{grow\_until\_certify} algorithm.

\section{Mathematical Proofs for CAGE-NAS}
\label{app:selected-cage-proofs}
\subsection{Gradient, smoothness, and PL identity of the code loss}
\label{app:fgd-mse-properties}
\paragraph{Proposition.}
The code loss is the summed squared error,
\[
 \mathcal{L}(F)=\|F-Y\|_2^2,
 \qquad F,Y\in\mathbb{R}^{s},
\]
and is twice Fréchet differentiable and satisfies
\begin{equation}
 r(F):=\nabla_F\mathcal{L}(F)=2(F-Y),
 \qquad
 \nabla_F^2\mathcal{L}(F)=2I_s.
 \label{eq:app-fgd-mse-gradient-hessian}
\end{equation}
Its exact smoothness and strong-convexity constants are \(L_s=2\) and
\(\mu=2\); its unique minimizer is \(F^\star=Y\), with
\(\mathcal{L}^\star=0\), and the PL inequality is attained as an equality.
\paragraph{Proof.}
For an arbitrary perturbation \(H\in\mathbb{R}^{s}\), expansion of the
squared norm gives
\begin{equation}
 \mathcal{L}(F+H)
 =
 \mathcal{L}(F)
 +2\langle F-Y,H\rangle
 +\|H\|_2^2.
 \label{eq:app-fgd-mse-exact-increment}
\end{equation}
The linear part is
\(A_F[H]=2\langle F-Y,H\rangle\), whereas the remainder is
\(R_F(H)=\|H\|_2^2\).  Since
\[
 \frac{|R_F(H)|}{\|H\|_2}
 =
 \|H\|_2
 \longrightarrow0
 \quad\text{as }\|H\|_2\longrightarrow0,
\]
\(\mathcal{L}\) is Fréchet differentiable and
\[
 D\mathcal{L}(F)[H]
 =
 2\langle F-Y,H\rangle.
\]
The defining relation
\(D\mathcal{L}(F)[H]
=\langle\nabla_F\mathcal{L}(F),H\rangle\)
therefore yields
\(\nabla_F\mathcal{L}(F)=2(F-Y)\).  Differentiating this affine gradient gives
\(\nabla_F^2\mathcal{L}(F)=2I_s\).
For any \(F,G\in\mathbb{R}^{s}\),
\[
 \|\nabla_F\mathcal{L}(F)-\nabla_F\mathcal{L}(G)\|_2
 =
 2\|F-G\|_2,
\]
so the exact global smoothness constant is \(L_s=2\).  Moreover,
\(\nabla_F^2\mathcal{L}=2I_s\succeq2I_s\), proving
\(2\)-strong convexity.  The only stationary point satisfies
\(2(F-Y)=0\); hence
\[
 F^\star=Y,
 \qquad
 \mathcal{L}^\star=0.
\]
Finally,
\begin{equation}
 \|\nabla_F\mathcal{L}(F)\|_2^2
 =
 4\|F-Y\|_2^2
 =
 4\bigl(\mathcal{L}(F)-\mathcal{L}^\star\bigr).
 \label{eq:app-fgd-mse-pl-identity}
\end{equation}
Thus
\[
 \mathcal{L}(F)-\mathcal{L}^\star
 =
 \frac{1}{2\mu}\|\nabla_F\mathcal{L}(F)\|_2^2
 \quad\text{with}\quad
 \mu=2.
\]
The constants \(L_s=\mu=2\) follow from the code-loss normalization; dividing
the loss by \(n\) would rescale both constants by \(1/n\). \(\square\)
\paragraph{Hilbert specialization of \(\alpha\) and \(\beta\).}
The general framework of \citet{csillag2026functional} distinguishes the
Hilbert space \(\mathcal{H}\), in which the functional gradient is defined,
from a Banach space \(\mathcal{B}\), whose norm controls smoothness and the
approximation error.  Writing
\[
 r=\nabla_F\mathcal L(F),
\]
the gradient-compatibility hypotheses introduce constants
\(\alpha,\beta>0\) satisfying
\begin{equation}
 D\mathcal L(F)[r]
 \geq
 \alpha\|r\|_{\mathcal{B}}^{2},
 \qquad
 \|D\mathcal L(F)\|_{\mathcal{B}^\ast}
 \leq
 \beta\|r\|_{\mathcal{B}}.
 \label{eq:fgd-alpha-beta-general}
\end{equation}
For the empirical problem executed by the code, both geometries are the
same:
\[
 \mathcal{B}=\mathcal{H}=\mathcal{H}_n\simeq\mathbb R^{s},
\]
with the Euclidean inner product and its induced norm.  By the defining
identity of the Hilbert gradient,
\[
 D\mathcal L(F)[H]
 =
 \langle r,H\rangle_{\mathcal{H}}
 \qquad\text{for every }H\in\mathcal{H}.
\]
Choosing \(H=r\) gives
\[
 D\mathcal L(F)[r]
 =
 \langle r,r\rangle_{\mathcal{H}}
 =
 \|r\|_{\mathcal{H}}^{2}.
\]
Moreover, the Riesz representation theorem identifies the functional
\(D\mathcal L(F)\in\mathcal{H}^\ast\) with its unique representer \(r\).
This identification is isometric:
\[
 \|D\mathcal L(F)\|_{\mathcal{H}^\ast}
 =
 \sup_{\|H\|_{\mathcal{H}}=1}
 |\langle r,H\rangle_{\mathcal{H}}|
 =
 \|r\|_{\mathcal{H}},
\]
where Cauchy--Schwarz gives the upper bound and equality is attained at
\(H=r/\|r\|_{\mathcal{H}}\) when \(r\neq0\) (the case \(r=0\) is immediate).
Consequently, both inequalities in
\eqref{eq:fgd-alpha-beta-general} hold as equalities for
\begin{equation}
 \alpha=1,\qquad\beta=1.
 \label{eq:fgd-hilbert-alpha-beta}
\end{equation}
\subsection{Sufficient descent and the implemented rate interval}
\label{app:fgd-lemma35-proof}
We specialize the sufficient-descent argument of
\citet{csillag2026functional} to the finite-dimensional Hilbert space used
by the program.
\paragraph{Proposition.}
Let \(r=\nabla_F\mathcal{L}(F)\), let \(g\neq0\), and suppose
\(\epsilon=\|g-r\|_B/\|g\|_B<1/2\), as in
Equation~\eqref{eq:fgd-relative-error-code}. For an \(L_s\)-smooth
functional, the sufficient-descent inequality is
Equation~\eqref{eq:fgd-code-descent}, with
\[
 q(\eta,\epsilon)
 =
 1-\frac{L_s\eta}{2}
 -\left(1+\frac{3L_s\eta}{2}\right)
 \frac{\epsilon}{1-\epsilon}.
\]
In particular, \(q>0\) on the rate interval in
Equation~\eqref{eq:fgd-rate-ceiling}.
\paragraph{Specialization.}
The displayed descent inequality is Lemma~3.5 of
\citet{csillag2026functional}, so it is not reproved here.  Substituting the
Hilbert values \(\alpha=\beta=1\) and the code-loss smoothness constant
\(K=L_s=2\) into that lemma gives the stated coefficient \(q\).  The same
result requires this coefficient to be strictly positive; solving
\(q(\eta,\epsilon)>0\) gives
\begin{align}
 1-\frac{L_s\eta}{2}
 -\left(1+\frac{3L_s\eta}{2}\right)
 \frac{\epsilon}{1-\epsilon}
 &>0 \nonumber\\
 (1-\epsilon)
 -\frac{L_s\eta}{2}(1-\epsilon)
 -\epsilon
 -\frac{3L_s\eta}{2}\epsilon
 &>0 \nonumber\\
 1-2\epsilon
 -\frac{L_s\eta}{2}(1+2\epsilon)
 &>0 \nonumber\\
 \frac{L_s\eta}{2}(1+2\epsilon)
 &<1-2\epsilon.
 \label{eq:app-fgd-rate-bound-isolation}
\end{align}
Because \(L_s>0\) and \(0\leq\epsilon<1/2\), division by
\(\frac{L_s}{2}(1+2\epsilon)>0\) preserves the inequality and yields
exactly Equation~\eqref{eq:fgd-rate-ceiling}; for \(L_s=2\), its upper
bound is \((1-2\epsilon)/(1+2\epsilon)\). \(\square\)
\subsection{Global linear contraction for the implemented training loss}
\label{app:fgd-global-contraction}
Combining the sufficient-descent inequality with
\(r_t:=r(F_t)\), \(\|r_t\|_B^2=4\mathcal{L}(F_t)\), and
\(q_t:=q(\eta_t,\epsilon_t)\) yields
\begin{align}
 \mathcal{L}(F_{t+1})
 &\leq
 \mathcal{L}(F_t)
 -4\eta_tq_t\mathcal{L}(F_t) \nonumber\\
 &=
 (1-4\eta_tq_t)\mathcal{L}(F_t).
 \label{eq:app-fgd-one-step-contraction}
\end{align}
Thus, for committed steps with positive \(q_t\),
\begin{equation}
 \mathcal{L}(F_T)
 \leq
 \left[
 \prod_{t=0}^{T-1}(1-4\eta_tq_t)
 \right]
 \mathcal{L}(F_0).
 \label{eq:app-fgd-global-contraction}
\end{equation}
This is Proposition~3.8 with \(\mu=2\) and \(\beta=1\).  In the main
configuration the accumulated stationary and global bounds are logged as
trajectory diagnostics; the paper-pure
\texttt{certify\_apply\_in\_\allowbreak interval} path commits the certified training step from
its own relative-error certificate rather than demanding a second
held-out-loss gate.

\section{The damped tangent family}
\label{app:damped-tangent-family}
From the analysis presented by \cite{rudkiewicz2024fogro}, it follows naturally
that one of the first possibilities for defining the approximation family
arises from the tangent space in
Equation~\eqref{eq:demeter-fgd-tangent-space}. This space ultimately defines
the directions along which infinitesimal variations can be produced for a
given network architecture.
Therefore, as a first approximation, we consider the projection of the
functional gradient onto the tangent space, regularized by a damping factor
\(\lambda\geq 0\). More precisely, it solves the Tikhonov problem
\begin{equation}
    u_\lambda
    =
    \arg\min_{u\in\mathbb{R}^{p_A}}
    \left\{
        \|J_{A,\theta}u-r\|_2^2
        +
        \lambda\|u\|_2^2
    \right\},
    \qquad
    g_\lambda
    =
    J_{A,\theta}u_\lambda.
    \label{eq:fgd-damped-projection}
\end{equation}
In this formulation, the first term,
\(\|J_{A,\theta}u-r\|_2^2\), measures how closely the contribution of a
direction in parameter space to the neural network output approximates the
functional gradient. The second term is introduced to regularize the
projection and discourage large parameter movements.
This regularization is necessary because two parameter directions can have
nearly the same first-order effect on the output while requiring radically
different parameter norms. A very large \(u\) can increase the neglected
nonlinear Taylor remainder and invalidate the translation from parameter
space to function space, even when
\(\|J_{A,\theta}u-r\|_2\) is small.
Since the empirical representation space in which the neural network is
defined is a finite-dimensional Hilbert space, the empirical Jacobian is a
finite matrix,
\[
    J_{A,\theta}
    =
    D\Phi_A(\theta)
    \in
    \mathbb{R}^{s\times p_A}.
\]
This matrix can be factorized through a singular value decomposition as
\[
    J_{A,\theta}
    =
    U\Sigma V^\top,
    \qquad
    a
    =
    U^\top r,
\]
where \(\sigma_i\) are the singular values. From this representation, it is
possible to obtain the direction in parameter space that best supports the
step in the functional gradient descent direction, together with the
functional-gradient approximation at the neural network output. The latter
corresponds to the final functional contribution of the parameter update in
the computed direction:
\begin{align}
    u_\lambda
    &=
    V\operatorname{diag}
    \left(
        \frac{\sigma_i}{\sigma_i^2+\lambda}
    \right)a,
    \label{eq:fgd-spectral-update}\\
    g_\lambda
    &=
    U\operatorname{diag}
    \left(
        \frac{\sigma_i^2}{\sigma_i^2+\lambda}
    \right)a.
    \label{eq:fgd-spectral-function-direction}
\end{align}
These equations describe the transformation mode by mode. The vector
\(U^\top r\) measures how much of the desired gradient lies along each left
singular direction of the Jacobian. The factor
\begin{equation}
    h_i(\lambda)
    =
    \frac{\sigma_i^2}{\sigma_i^2+\lambda}
    \in
    [0,1]
    \label{eq:fgd-output-transmission-coefficient}
\end{equation}
is the corresponding output-space transmission coefficient:
\begin{itemize}
    \item If \(\sigma_i^2\gg\lambda\), then
    \(h_i(\lambda)\approx 1\), and that component of \(r\) is almost fully
    reproduced.
    \item If \(\sigma_i^2\ll\lambda\), then
    \(h_i(\lambda)\approx 0\), and the poorly conditioned component is
    suppressed.
    \item If \(\sigma_i=0\), the component is structurally unreachable by any
    infinitesimal parameter change, and no damping value can recover it.
\end{itemize}
The parameter filter
\[
    \frac{\sigma_i}{\sigma_i^2+\lambda}
\]
behaves differently. Without damping, it becomes \(1/\sigma_i\), so weak
singular modes can produce extremely large parameter updates. Damping
therefore trades functional fidelity for a smaller and more realizable
parameter displacement.
The approximation certificate follows the nonstandard normalization used by
\citet{csillag2026functional}:
\begin{equation}
 \epsilon_\lambda
 =
 \operatorname{RelErr}(g_\lambda,r)
 =
 \frac{\|g_\lambda-r\|_B}{\|g_\lambda\|_B}
 =
 \frac{
  \|v_{\lambda}^{\perp,\mathrm{damp}}\|_{L^2}
 }{
  \|v_{\lambda}^{\star,\mathrm{damp}}\|_{L^2}
 }.
 \label{eq:app-fgd-tangent-relative-error}
\end{equation}
It can also be observed in
Equation~\eqref{eq:app-fgd-tangent-relative-error} that, when the approximation family
is considered within the tangent space induced by the network architecture,
the relative error of the approximation function retains the expressivity
bottleneck in the numerator.
This indicates that the approximation is a good approximation in the sense of
functional gradient descent when the norm of the expressivity bottleneck is
smaller than the norm of the tangent-function approximation. This implies that
the greater the expressive capacity of the neural network architecture, the
smaller the expressivity bottleneck and, consequently, the smaller the
relative error.
This approximate descent direction is admissible only when the condition
\begin{equation}
    \epsilon_\lambda
    <
    \min
    \left\{
        \tau_{\mathrm{rel}},
        \frac{1}{2}
    \right\}
    =
    \frac{1}{2}
    \label{eq:fgd-admissibility}
\end{equation}
holds, where \(\tau_{\mathrm{rel}}\) denotes the configured relative-error
acceptance threshold, that is, the minimum approximation quality required
for admitting the direction. This denominator must not be replaced by
\(\|r\|_B\): the proof, the
threshold, and the implementation all use \(\|g_\lambda\|_B\).
In the undamped orthogonal-projection limit, write
\[
    r
    =
    r_{\parallel}
    +
    r_{\perp},
    \qquad
    r_{\parallel}
    \in
    T_{A,\theta},
    \qquad
    r_{\perp}
    \perp
    T_{A,\theta}.
\]
Then \(g=r_{\parallel}\), and
\begin{equation}
    \epsilon^2
    =
    \frac{
        \|r_{\perp}\|_B^2
    }{
        \|r_{\parallel}\|_B^2
    }.
    \label{eq:fgd-bottleneck-ratio}
\end{equation}
Therefore, \(\epsilon\) is a direct ratio between the unreachable energy
and the reachable signal. The condition \(\epsilon<1/2\) implies
\begin{equation}
    \|r_{\parallel}\|_B^2
    >
    \frac{1}{
        1+(1/2)^2
    }
    \|r\|_B^2
    =
    0.8\|r\|_B^2.
    \label{eq:fgd-eighty-percent-energy}
\end{equation}
Geometrically, the current tangent space must capture more than \(80\%\) of
the functional-gradient energy. With damping, \(g_\lambda\) is a filtered
projection rather than an orthogonal projection. Therefore, this \(80\%\)
identity should be understood as an interpretive limiting case, while the
implemented relative error is evaluated directly.
Using the Hilbert-space values \(\alpha=\beta=1\) proved in
Appendix~\ref{app:fgd-mse-properties}, together with \(L_s=2\), the expression
describing the sufficient descent produced by each functional gradient descent
step follows from \citet{csillag2026functional}; its specialization to the
implemented constants is recorded in
Appendix~\ref{app:fgd-lemma35-proof}:
\begin{equation}
    \mathcal{L}(F-\eta g_\lambda)
    \leq
    \mathcal{L}(F)
    -
    \eta\,
    q(\eta,\epsilon_\lambda)
    \|r\|_B^2,
    \label{eq:app-fgd-tangent-descent}
\end{equation}
where
\begin{equation}
    q(\eta,\epsilon)
    =
    1-\eta
    -
    (1+3\eta)
    \frac{\epsilon}{1-\epsilon}.
    \label{eq:fgd-code-descent-coefficient}
\end{equation}
Therefore, the learning rate applied at each step must guarantee that
\(q(\eta,\epsilon)\) is strictly positive, so that the descent reduces the
loss by an amount proportional to the current value of the gradient norm.
Consequently, the admissible learning rate is subject to the upper bound
\begin{equation}
    \bar{\eta}(\epsilon)
    =
    \frac{2(1-2\epsilon)}
    {L_s(1+2\epsilon)}
    =
    \frac{1-2\epsilon}
    {1+2\epsilon}.
    \label{eq:app-fgd-tangent-rate-ceiling}
\end{equation}
An additional lower bound is imposed to guarantee that steps with excessively
small learning rates, which would produce almost no variation and could
therefore cause stagnation of the loss function despite satisfying the minimum
descent criterion, are not performed:
\begin{equation}
    \eta_{\mathrm{cert}}
    =
    \frac{1}{2}
    \bar{\eta}(\epsilon_\lambda),
    \qquad
    \eta_{\mathrm{cert}}
    >
    10^{-5}.
    \label{eq:fgd-optimal-certified-rate}
\end{equation}
For the application of the algorithm, we propose selecting the midpoint of the
admissible interval, which maximizes the quadratic lower bound on the guaranteed
descent. The rate decreases continuously as the approximation deteriorates.
With a perfect direction, the step uses \(\eta=1/2\) and reaches \(Y\).
As the direction becomes less trustworthy, the permitted movement collapses toward
zero. At \(\epsilon=1/2\), no strictly positive learning rate remains. This
is why the same value acts both as the admissibility threshold and as the
signal that the representation must be refined.
If, during the gradient descent stages, \(J_{A,\theta}\) is affected by a
transformation that scales it as
\[
    \widetilde{J}
    =
    \kappa J_{A,\theta},
\]
the tangent space remains unchanged. This necessarily implies that the
architecture can continue representing the same functional directions.
However, when \(\lambda\) is kept fixed, the filter becomes
\[
    \widetilde{h}_i(\lambda)
    =
    \frac{
        \kappa^2\sigma_i^2
    }{
        \kappa^2\sigma_i^2+\lambda
    },
\]
which demonstrates that a fixed \(\lambda\) is not scale-invariant. Therefore,
to preserve the balance between the data term and the regularization term
\(\lambda\|u\|^2\), an adaptive damping parameter is proposed:
\begin{equation}
    \lambda
    =
    \rho\,\sigma_{\max}^2,
    \qquad
    \rho
    \in
    [10^{-16},10^2].
    \label{eq:fgd-relative-damping}
\end{equation}
The dimensionless parameter \(\rho\) makes the choice relative to the largest
local sensitivity of the network. If all Jacobian singular values are
multiplied by \(\kappa\), then \(\sigma_{\max}^2\) and \(\lambda\) are both
multiplied by \(\kappa^2\), leaving every filter
\[
    \frac{
        \kappa^2\sigma_i^2
    }{
        \kappa^2\sigma_i^2
        +
        \rho\kappa^2\sigma_{\max}^2
    }
    =
    \frac{
        \sigma_i^2
    }{
        \sigma_i^2
        +
        \rho\sigma_{\max}^2
    }
\]
unchanged. The search is therefore invariant to a global rescaling of the
Jacobian spectrum.
Since every spectral filter in the preceding scale-invariant expression
decreases with \(\rho\), \(\epsilon_\lambda\) is searched from the
least-regularized end. The implementation first performs \(40\) geometric
bisection steps to locate the largest \(\rho_\star\) satisfying
\(\epsilon_{\rho_\star}<1/2\). It then evaluates \(25\) rungs
\begin{equation}
    \rho_k
    =
    \rho_\star\,10^{-k/4},
    \qquad
    k
    =
    0,\ldots,24,
    \label{eq:fgd-damping-fan}
\end{equation}
reusing the same SVD. Among the rungs that both satisfy the certification
condition and produce a finite realizable rate, the active objective
\texttt{descent} selects
\begin{equation}
    \rho^{\star}
    \in
    \arg\max_{\rho_k}
    \eta_k
    \|g_{\lambda_k}\|_B^2.
    \label{eq:fgd-damping-objective}
\end{equation}
This is the spectral score. It favors a large realizable movement within the
tangent image, but it does not correspond exactly to the complete sufficient
descent lower bound
\[
    \eta_k
    q(\eta_k,\epsilon_k)
    \|r\|_B^2,
\]
because both \(q\) and \(\epsilon_k\) vary across the rungs.
The damping search is best understood as satisfying two opposing
requirements:
\begin{enumerate}
    \item \emph{Approximation}: decreasing \(\lambda\) transmits more singular
    modes and tends to reduce \(\epsilon_\lambda\).
    \item \emph{Realizability}: decreasing \(\lambda\) can amplify the
    parameter coefficients as \(1/\sigma_i\), increasing
    \(\|u_\lambda\|_2\) and the nonlinear Taylor remainder.
\end{enumerate}
A very large damping value produces a stable but uninformative direction,
whereas a very small damping value produces an accurate linear prediction that
may require an unrealistically large parameter update. The bisection identifies
the boundary at which certification remains possible, while the downward fan
explores less-regularized candidates on the admissible side of that boundary. Because all rungs share \(U\), \(\Sigma\), and \(V\), this exploration only
changes the spectral weights and does not modify the underlying derivative
calculation.

\section{Nonlinear parametric displacement family}
\label{app:nonlinear-parametric-family}
When the tangent approximation, due to the linearization inherent in the
first-order approximation of the effect of a parameter change on the loss, is
not sufficient, the search begins within the same network structure to
determine whether there exists a finite displacement direction.
This displacement can be understood as a parametric trajectory satisfying
\begin{equation}
    \Phi_A\bigl(\theta(T)\bigr)
    -
    \Phi_A\bigl(\theta(0)\bigr)
    =
    \int_{0}^{T}
    J_{A,\theta(t)}
    \dot{\theta}(t)
    \,\mathrm{d}t.
    \label{eq:fgd-finite-parametric-trajectory}
\end{equation}
Therefore, the network can reach a functional direction even when it does not
belong to the initial tangent space, by following a finite curved trajectory
composed of several locally realizable movements.
The parametric trajectory is constructed through training with gradient
descent, with the distinction that the target is not \(Y\). The objective is
not to directly seek the prediction, but rather to make the nonlinear
parametric displacement follow the functional-gradient trajectory.
Since \(r=2(F-Y)\) and \(\eta_f=1\),
\begin{equation}
    F_{\mathrm{target}}
    =
    F-2(F-Y)
    =
    2Y-F.
    \label{eq:fgd-family-reflected-target}
\end{equation}
This target is the reflection of \(F\) across \(Y\). The purpose of this inner
training is not to define \(2Y-F\) as the final network output. Instead, it is
to force a clone of the network to explore a finite and curved displacement
aligned with the functional descent ray.
Given the way in which the contribution is generated through the nonlinear
parametric trajectory, it is necessary to detail how the functional-gradient
descent approximation produced by the nonlinear family is certified.
The candidate functional direction is the realized nonlinear displacement
\begin{equation}
    \Delta
    =
    \Phi_A(\theta)
    -
    \Phi_A(\widetilde{\theta}).
    \label{eq:fgd-family-displacement}
\end{equation}
Because \(\Delta\) follows the curved parametric manifold rather than only its
tangent plane, it can contain second- and higher-order components. Let
\begin{equation}
    \chi=\frac{\langle\Delta,r\rangle}{\|\Delta\|_B\|r\|_B}.
    \label{eq:fgd-family-cosine-similarity}
\end{equation}
Along the ray \(g=a\Delta\), with \(a>0\), the smallest attainable
code-normalized relative error is
\begin{equation}
    \epsilon_{\mathrm{NL}}^\star=\sqrt{1-\max\{\chi,0\}^2}.
    \label{eq:fgd-family-cosine-error}
\end{equation}
The clone is committed only under the equivalent conditions
\begin{equation}
    \chi>0,\quad \epsilon_{\mathrm{NL}}^\star<\frac12
    \quad\Longleftrightarrow\quad
    \chi>\frac{\sqrt3}{2}.
    \label{eq:fgd-family-angular-admissibility}
\end{equation}
As becomes evident, for the certification of sufficient gradient descent in
the FGD step, what is evaluated in this family is the satisfaction of the
relative-error criterion in the direction of the contribution produced by the
parametric trajectory, rather than the magnitude of the contribution itself.
Because of the nonlinear nature of the displacement, the optimal scaling is
\begin{equation}
    a^\star=\frac{\|r\|_B^2}{\langle\Delta,r\rangle}.
    \label{eq:fgd-family-optimal-code-scaling}
\end{equation}
Although \(\Delta\) may have an appropriate orientation with respect to \(r\),
its magnitude depends on how far the parameters moved during the training of
the clone. This magnitude is affected by the clone learning rate, the number
of iterations, the damping, the curvature of \(\Phi_A(\theta)\), and the nonlinear
terms ignored by the tangent approximation.
Therefore, it is possible that
\[
    \Delta\parallel r,\qquad
    \|\Delta\|_B\gg\|r\|_B
    \quad\text{or}\quad
    \|\Delta\|_B\ll\|r\|_B.
\]
The scaling \(a^\star\) is intended precisely to correct this difference in
magnitude, such that
\[
    a^\star\Delta
    \approx
    r.
\]

\section{Controlled results, benchmark, and grid searches}
\label{app:controlled-results}

\subsection{Certification, Growth, and Performance: An Empirical Study of CAGE-NAS}

As has been extensively evidenced, the computational cost associated with the
tangent operations represents a significant limitation in the performance of
the CAGE-NAS method. For this reason, in order to obtain a theoretical
certification of the behavior of the algorithm, we initially propose validating
its theoretical formulation in a controlled environment with a small dataset
that can be used in full for certification and in which the dense
algorithm can be executed with the exact computation of the quantities
associated with the tangent space, as detailed in
Appendix~\ref{app:synthetic-n1024}.

The results in this controlled environment are obtained using the dense
execution path of the algorithm, with stopping determined by a parameter
budget for the neural network architecture. Additionally, given the nature of
the algorithm, we propose starting from a small architecture composed of three
hidden layers with two neurons in each layer. The experiment is repeated over four random seeds to evaluate the
reproducibility of the algorithm. This is particularly relevant because, given
the reduced size of the initial network architecture, the resulting dynamics
are highly likely to depend on the initialization and may therefore affect the
final result.

All controlled \(N=1024\) CAGE-NAS train-and-grow results reported in this
subsection (Algorithms~\ref{alg:epoch-level-certified-training}
and~\ref{alg:certified-outer-step}) use the
\texttt{family\_ladder\_N1024} configuration with
\texttt{certify\_apply\_in\_interval=true}. Each tangent or nonlinear-family
functional direction is certified exclusively on the fixed certification set containing
all \(16\times64=1024\) training samples. After the certified rate is selected
and its functional target is realized, the step is committed without an
additional validation- or test-loss acceptance gate. Validation and test
losses are reported as held-out trajectory and generalization diagnostics and
do not enter this functional-step acceptance decision; consequently, the
held-out-loss gate represented by \textsc{EvaluateOuterStepAcceptance} is not
the path that produced the \(N=1024\) tables and figures.

Results show a favorable dynamic in which the model progressively adds
capacity during training while reducing the training loss at each epoch.
Additionally, the damping strategy contributes to the generalization of the
model. Another important result is that, although the algorithm depends considerably
on the initialization due to the small size of the initial network, and this
initialization significantly affects the complete growth-and-training
trajectory, the method remains able to reach satisfactory final solutions.
Evaluation loss is the mean squared error (MSE), and the quantitative
comparisons below report \(\operatorname{RMSE}=\sqrt{\operatorname{MSE}}\) in
the target's original units. The legacy accuracy panels retained in this
appendix are trajectory diagnostics: for this single-output regression task,
accuracy is the fraction of examples satisfying
\(\lvert\hat y_i-y_i\rvert\leq0.1\). It is not classification accuracy and is
not used for the RMSE rankings below.

The final architectures and test RMSE values of these four train-and-grow runs
are summarized in Table~\ref{tab:cage-grow-seed-results}; together, they reach
a mean test RMSE of \(0.0613\) with a standard deviation of \(0.0060\).

\begin{table}[H]
    \centering
    \caption{Final architectures and test RMSE values of the CAGE-NAS
    train-and-grow runs.}
    \label{tab:cage-grow-seed-results}
    \begin{tabular}{ccc}
        \hline
        \textbf{Seed} & \textbf{Architecture} & \textbf{Grow test RMSE} \\
        \hline
        0 & \([16,19,9]\)  & 0.0642 \\
        1 & \([9,18,18]\)  & 0.0680 \\
        2 & \([13,16,15]\) & 0.0585 \\
        3 & \([11,17,17]\) & 0.0544 \\
        \hline
    \end{tabular}
\end{table}

\begin{figure}[H]
    \centering
    \subfloat[Training loss.\label{fig:cage-n1024-train-loss}]{%
        \includegraphics[width=0.43\textwidth]{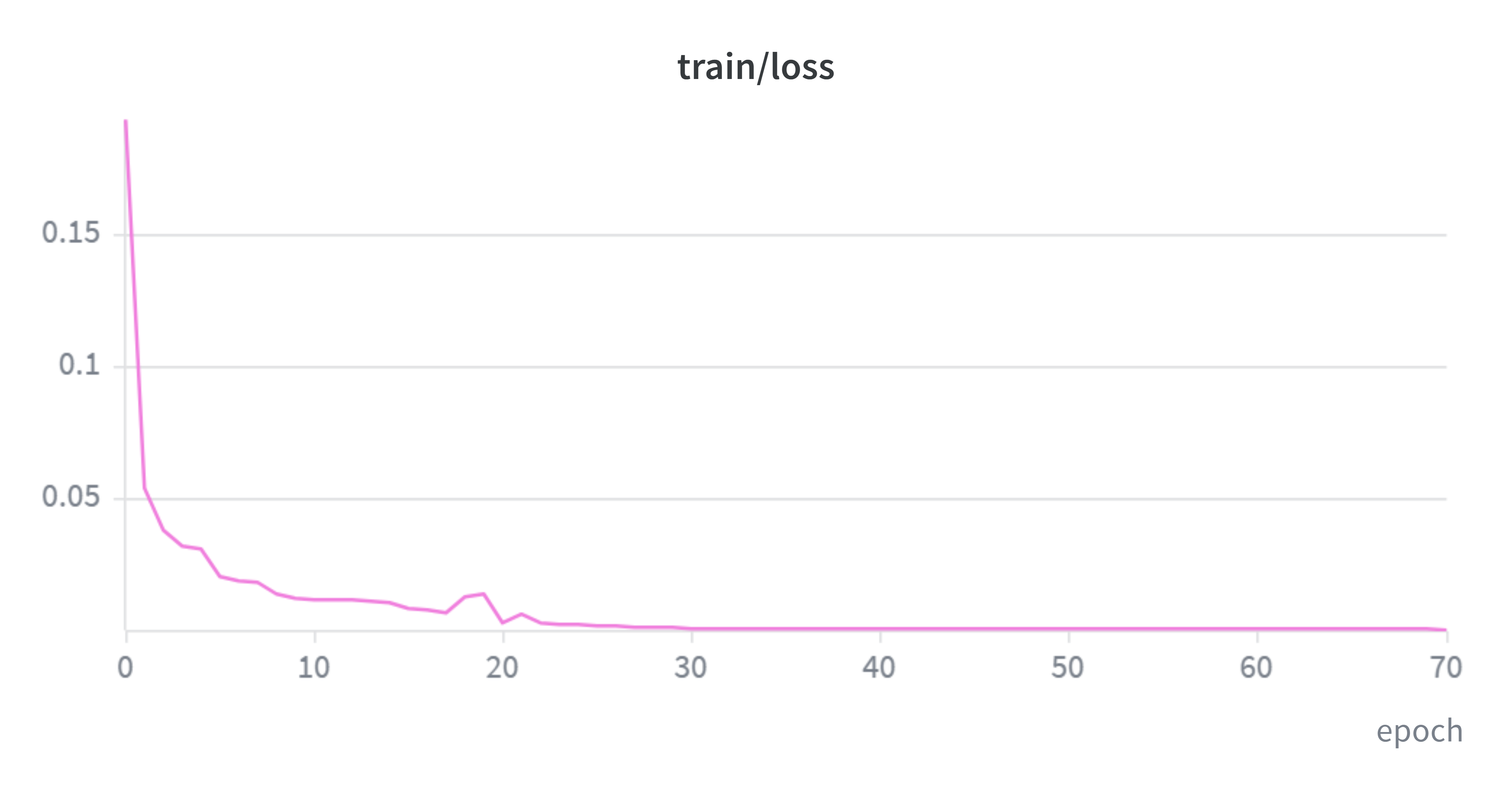}%
    }\hfill
    \subfloat[Training accuracy.\label{fig:cage-n1024-train-accuracy}]{%
        \includegraphics[width=0.43\textwidth]{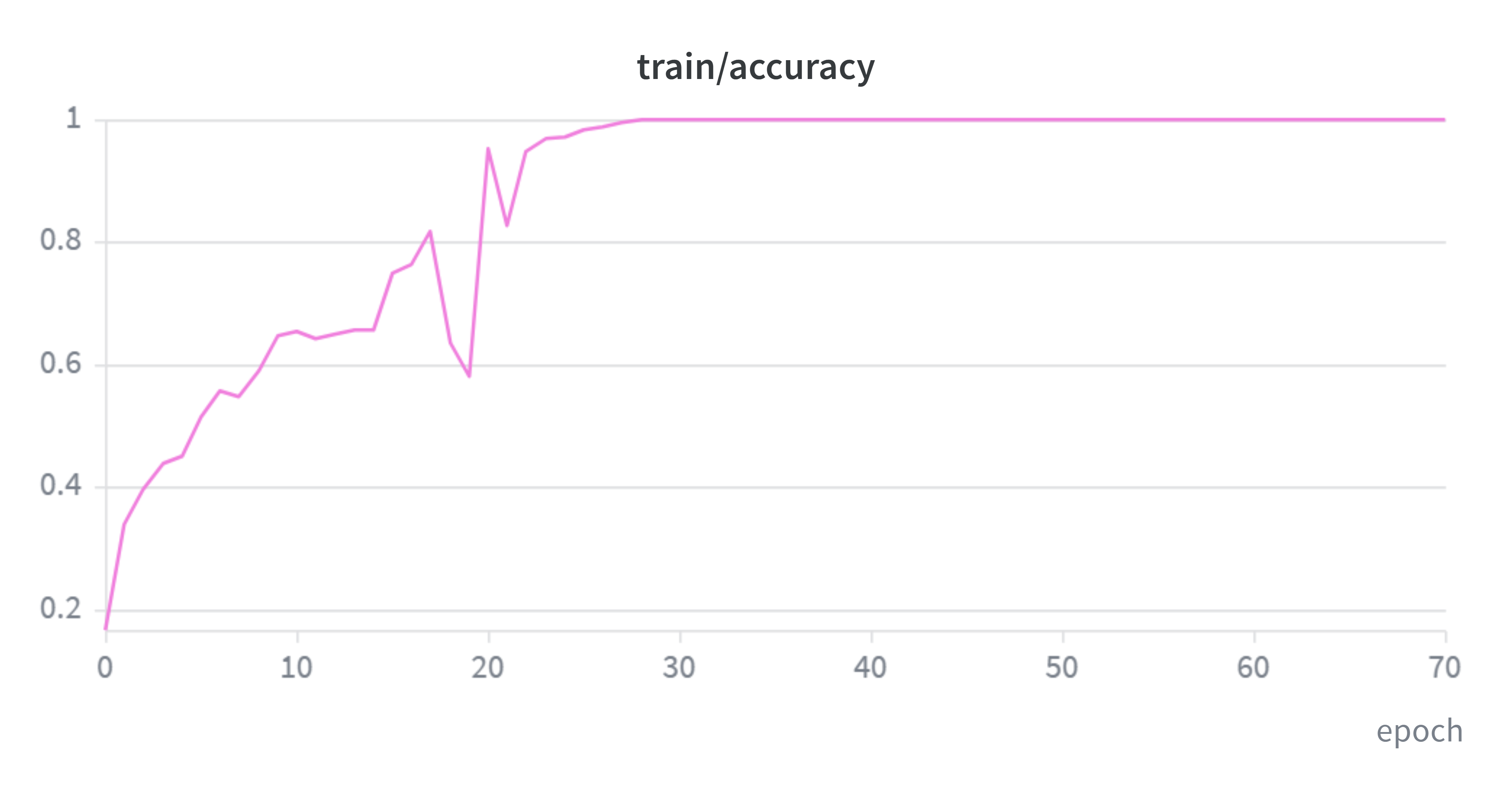}%
    }\\[0.2em]
    \subfloat[Number of parameters.\label{fig:cage-n1024-train-parameters}]{%
        \includegraphics[width=0.43\textwidth]{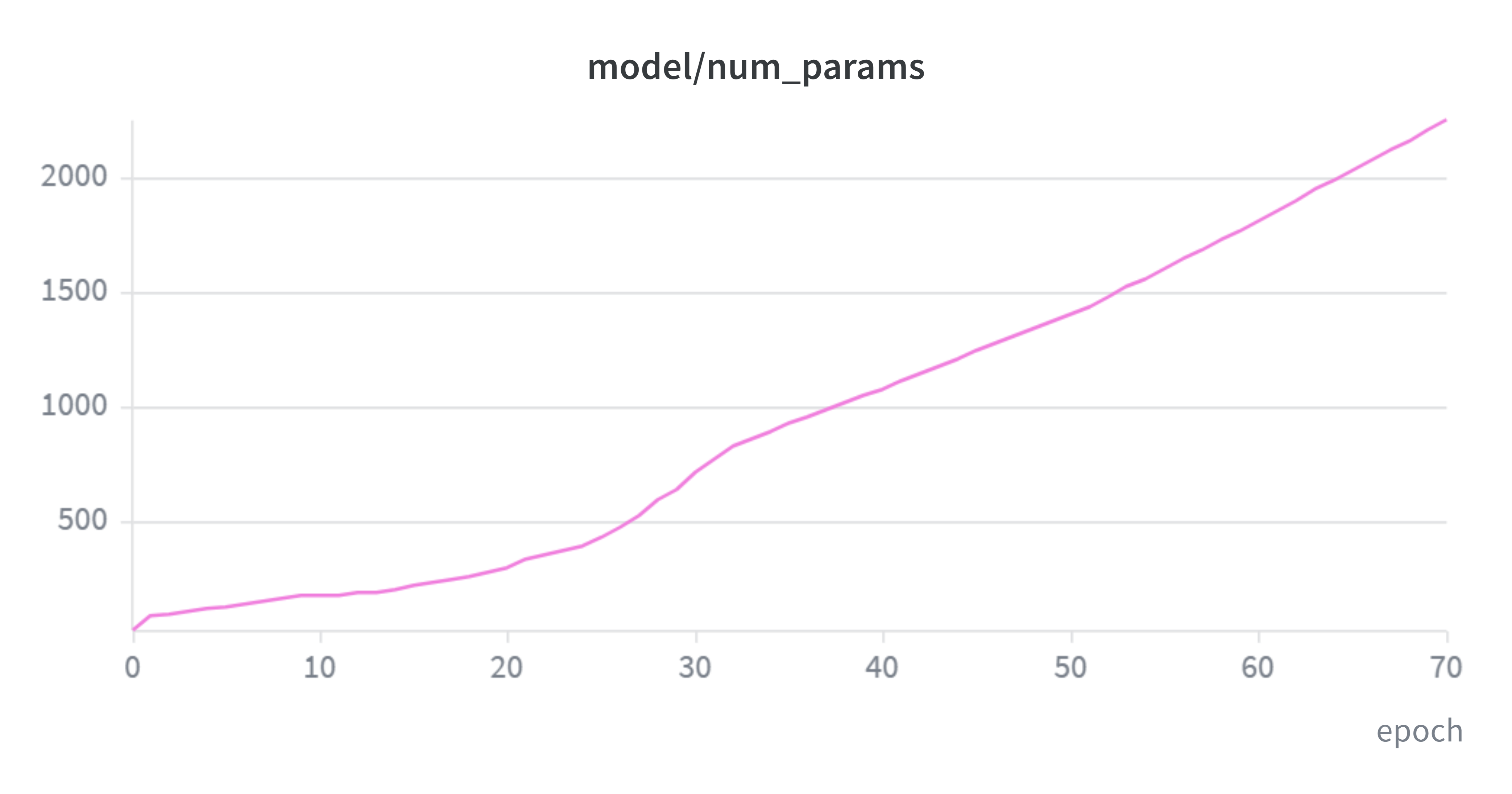}%
    }
    \caption{Training and architectural-growth behavior of CAGE-NAS on the
    controlled synthetic \(N=1024\) benchmark for seed \(0\) over \(70\)
    epochs.}
    \label{fig:cage-n1024-train-results}
\end{figure}

This behavior is illustrated in
Figure~\ref{fig:cage-n1024-train-results} for the representative
\(70\)-epoch run with seed \(0\). In addition to the loss reduction described
above, the training accuracy approaches one without persistent degradation
after the architecture is expanded. This indicates that the added capacity is
integrated stably and converted into optimization progress rather than merely
increasing the size of the model.

\begin{figure}[H]
    \centering
    \subfloat[Test loss.\label{fig:cage-n1024-test-loss}]{%
        \includegraphics[width=0.43\textwidth]{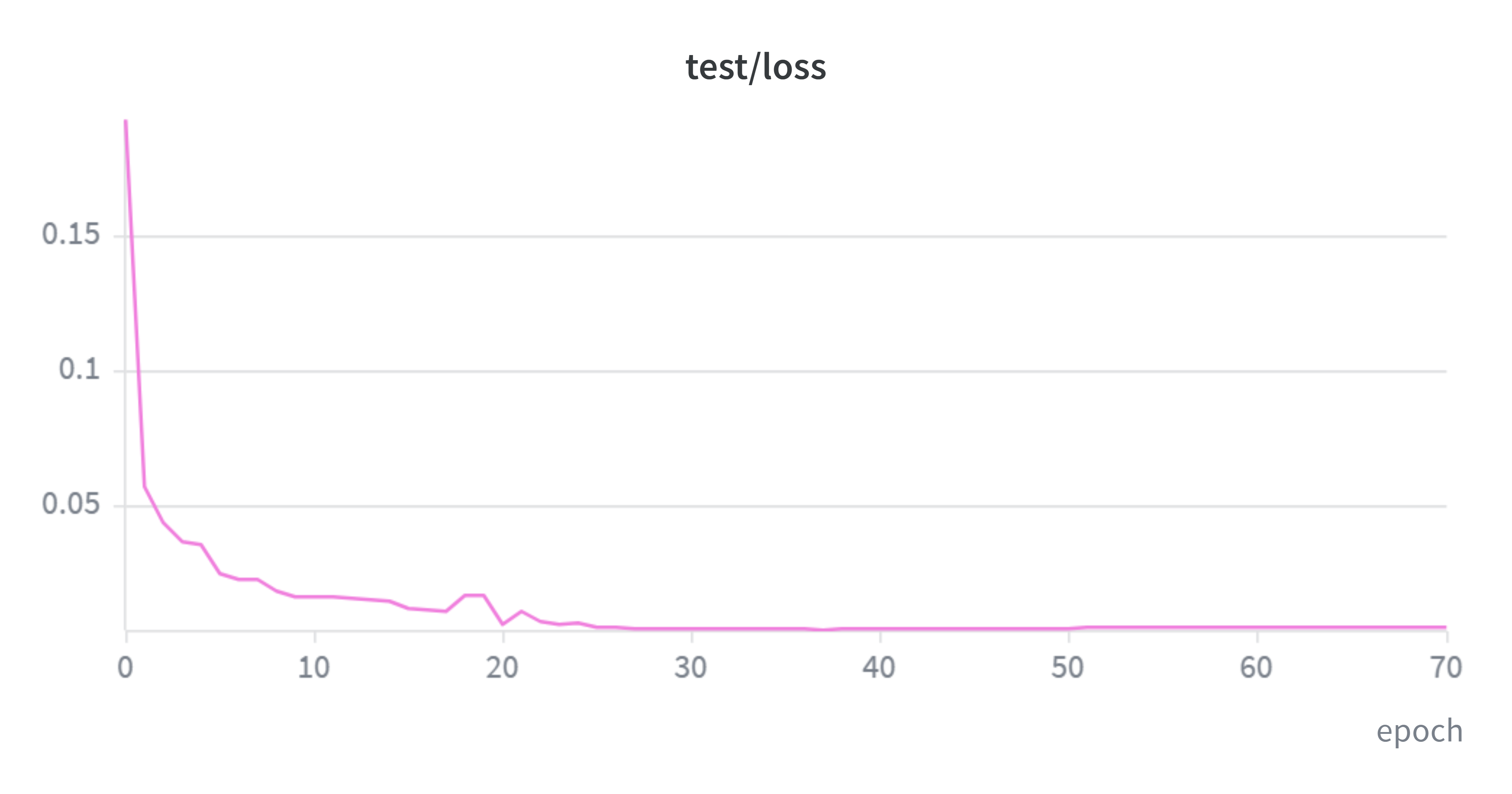}%
    }\hfill
    \subfloat[Test accuracy.\label{fig:cage-n1024-test-accuracy}]{%
        \includegraphics[width=0.43\textwidth]{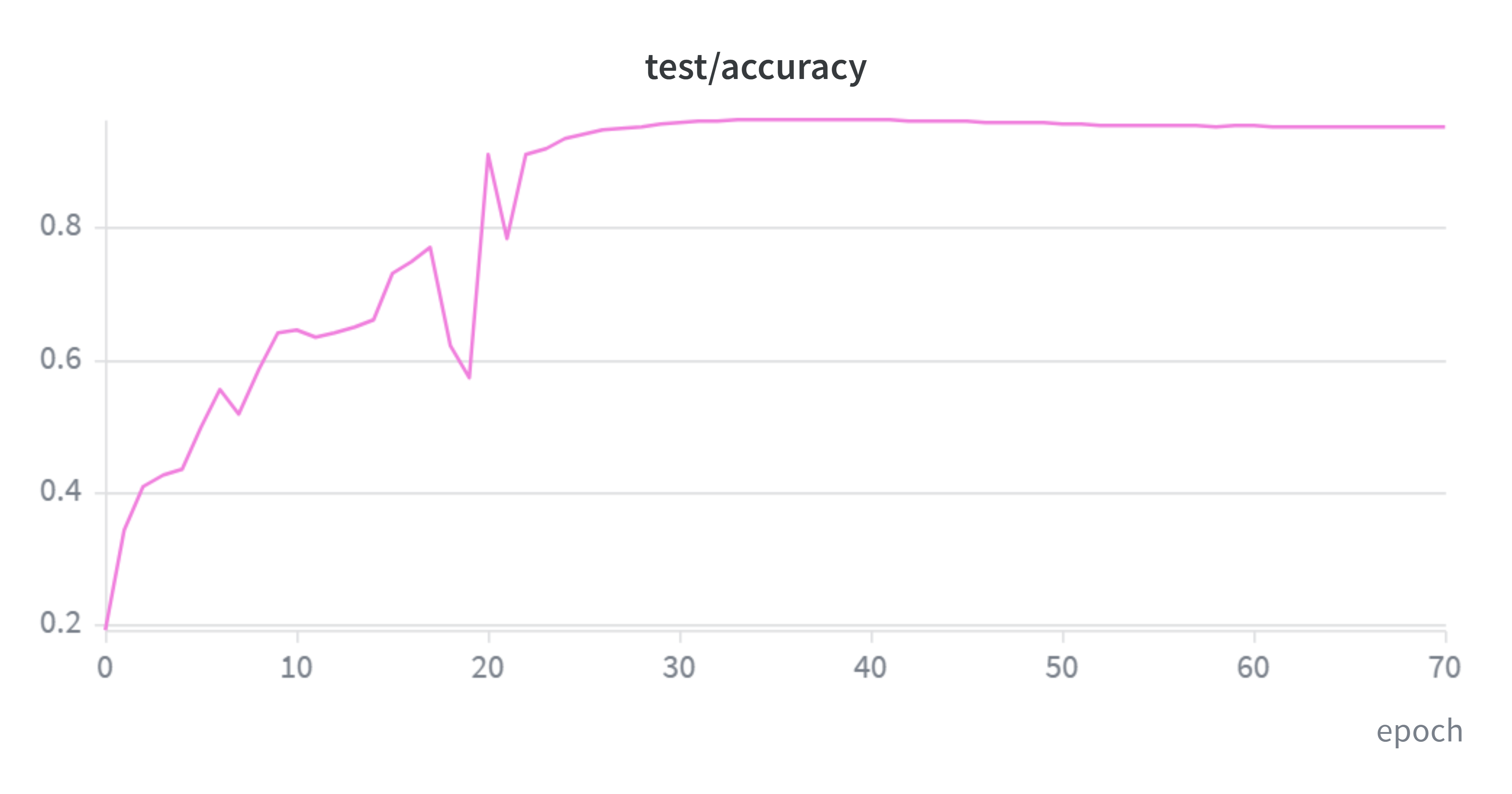}%
    }
    \caption{Test behavior of CAGE-NAS on the controlled synthetic \(N=1024\)
    benchmark for seed \(0\) over \(70\) epochs.}
    \label{fig:cage-n1024-test-results}
\end{figure}

The generalization effect mentioned above is reflected in the held-out curves
in Figure~\ref{fig:cage-n1024-test-results}. After a short initial transient,
the test loss follows a sustained downward trend while the test accuracy rises
and stabilizes near its upper plateau, without diverging from the training
trajectory. This agreement as the architecture grows shows that the additional
capacity is useful beyond the training data used for certification and provides empirical
evidence that the algorithm works effectively on this controlled \(N=1024\)
dataset.

Even for less favorable initializations, such as seed \(0\), the algorithm
requires a larger number of epochs to identify an adequate architecture under
the imposed parameter budget, but the final test RMSE remains stable across
the different runs. This supports the
repeatability of the method under different initializations. This observation also highlights a second important aspect: a stopping
criterion based only on execution time or number of epochs is insufficient.
The adaptive development of the training-and-growth trajectory can require a
larger number of epochs when the initialization is less favorable with respect
to the parameter budget allowed for the growing architecture.
Following the objective of the proposed algorithm, the goal is for both the
identified architecture and the training process performed during execution to
be optimal, or at least close to optimality. In this context, we compare the
final performance obtained by CAGE-NAS with the best results found for
admissible architectures within the corresponding parameter budget through the grid search presented in
Appendix~\ref{app:smaller-architecture-grid}.

The results do not establish that the algorithm identifies the globally best
architecture under the imposed parameter budget, since the grid search
contains individual candidates with lower test RMSE, as shown in
Table~\ref{tab:cage-grid-best-candidates}. Instead, this comparison makes it
possible to evaluate how close the discovered architectures are to the best
observed alternatives under the same budget.

\begin{table}[h]
    \centering
    \caption{Comparison between the final CAGE-NAS growth result and the
    candidate selected by minimum validation MSE from the saved strict-budget
    grid results for each seed. Lower test RMSE is better.}
    \label{tab:cage-grid-best-candidates}
    \begin{tabular}{ccc}
        \hline
        \textbf{Seed} &
        \textbf{Growth RMSE} &
        \textbf{Grid RMSE} \\
        \hline
        0 & 0.0642 & 0.0538 \\
        1 & 0.0680 & 0.0674 \\
        2 & 0.0585 & 0.0589 \\
        3 & 0.0544 & 0.0641 \\
        \hline
    \end{tabular}
\end{table}

\begin{figure}[H]
    \centering
    \includegraphics[width=0.82\textwidth]{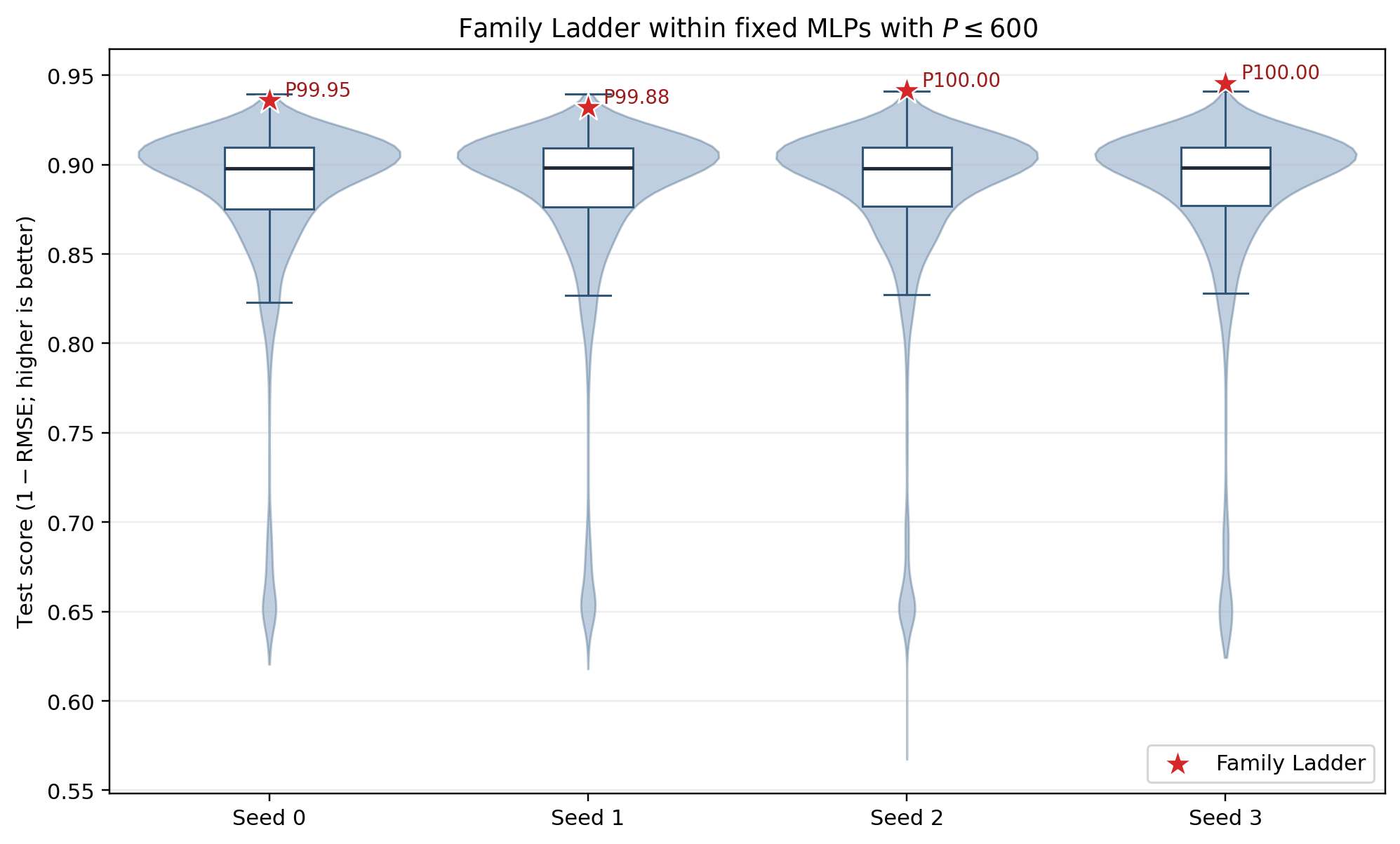}
    \caption{Distribution of the test score \(1-\mathrm{RMSE}\) for the screened
    fixed MLPs satisfying \(p(A)\leq600\). Red stars indicate the corresponding
    CAGE-NAS train-and-grow runs; higher is better.}
    \label{fig:cage-test-rmse-distribution}
\end{figure}

However, when analyzing the distribution of the results over all evaluated
architectures satisfying \(p(A)\leq600\), it can be observed that, in terms of
test RMSE, the model lies within the top \(0.2\%\) of the architectures found in the search for
every evaluated seed, as illustrated in
Figure~\ref{fig:cage-test-rmse-distribution}. Seed \(0\) reaches the
\(99.95\)th performance percentile and seed \(1\) the \(99.88\)th, while
seeds \(2\) and \(3\) reach the \(100.00\)th percentile. Here a higher
percentile denotes lower RMSE. These results do not
prove that the globally best architecture has been recovered, but they show
that CAGE-NAS consistently obtains architectures very close to the best
observed alternatives under the same parameter budget. The discovered models
lie in the extreme upper tail of a search space containing tens of thousands
of alternatives, without requiring those alternatives to be explicitly
enumerated during the train-and-grow trajectory.

Considering the computational effort involved, the budget-constrained alternatives
required screening \(76{,}646\) architectures and fine-tuning \(12{,}240\)
architecture--hyperparameter combinations for \(400\) epochs. Train-and-grow,
in contrast, generated each architecture through a single trajectory. Its
demonstrated advantage is therefore computationally efficient
architecture discovery.

\begin{table}[h]
    \centering
    \caption{Position of the CAGE-NAS growth result within the distribution of
    architectures evaluated by grid search.}
    \label{tab:cage-growth-percentiles}
    \begin{tabular}{ccc}
        \hline
        \textbf{Seed} &
        \textbf{Architectures evaluated} &
        \textbf{Growth percentile} \\
        \hline
        0 & 17{,}339 & 99.95 \\
        1 & 17{,}339 & 99.88 \\
        2 & 17{,}339 & 100.00 \\
        3 & 17{,}339 & 100.00 \\
        \hline
    \end{tabular}
\end{table}

These results highlight two key elements of fundamental analytical interest
within the behavior of CAGE-NAS. The first is the previously discussed strong dependence of the algorithm on
its initialization. Although this dependence still affects the final RMSE
and the exact position within the distribution, even the less favorable seed
\(1\) remains in the \(99.88\)th percentile. Initialization therefore changes
the relative ranking within the extreme upper tail rather than preventing the
algorithm from finding an architecture close to the best observed solutions
under the same parameter budget. The second point is that, although the
algorithm aims to provide convergence
and optimality guarantees for the functional-gradient trajectory, and this
behavior is reflected in the evolution of the training loss and training
error diagnostics, two additional aspects must be considered.

First, the formulation is defined as a sequence of mutually dependent events,
but the optimization path itself does not contain explicit information about
the number of parameters of the current architecture. Therefore, the number of
parameters is not directly optimized as part of the architectural search. This
may allow a smaller network to reach the same optimum obtained by the
successive growth steps, even though such a smaller architecture is not
explicitly favored by the optimization criterion. Second, although the optimum is pursued on the training objective, the
comparison presented here is performed using test data. Therefore, it is
possible that some architectures favor generalization more strongly but are
not selected during the train-and-grow process precisely because test
performance is unavailable during training. This suggests that an additional mechanism for favoring generalization could
be incorporated into the train-and-grow algorithm, allowing the method not only
to approach the training optimum but also to improve its test performance,
complementing the regularization effect already introduced through damping.

Another result of interest regarding the validation of the train-and-grow
sequence is the confirmation of whether, for the final architecture obtained,
freezing the architecture, reinitializing its neurons, and training it from
scratch using a PGD method improves the results obtained by the algorithm.
This is commonly the case when architectural search is performed using TINY
\cite{verbockhaven2024tiny} or its extension DEMETER.

To preserve the consideration of the initialization effect, the same seed used
during the growth algorithm is maintained when reinitializing the final
architecture. A grid search is then performed for the fixed architecture, as
documented in Appendix~\ref{app:fixed-architecture-grid}.

The paired results are mixed across individual seeds. Fixed retraining obtains
lower test RMSE for seed \(1\), whereas the train-and-grow
trajectory obtains lower test RMSE for seeds \(0\), \(2\), and \(3\). Thus,
growth wins three of the four paired comparisons when the architecture and
seed are held fixed.

Overall, fixed retraining achieves a mean test RMSE of \(0.0700\) with a
standard deviation of \(0.0053\), while growth achieves a mean of \(0.0613\)
with a standard deviation of \(0.0060\).

The paired difference, computed as fixed retraining minus growth, has a mean of
\(+0.0087\) with a standard deviation of \(0.0107\). Consequently, growth
obtains lower test RMSE on average, although the small four-seed sample does not
establish a uniform advantage.

The complete results are reported in
Table~\ref{tab:cage-fixed-vs-growth}.

\begin{table*}[t]
    \centering
    \small
    \setlength{\tabcolsep}{3pt}
    \caption{Comparison between the final CAGE-NAS train-and-grow result and
    retraining the discovered fixed architecture using a hyperparameter grid
    search.}
    \label{tab:cage-fixed-vs-growth}
    \begin{tabular}{cccccccc}
        \hline
        \textbf{Seed}
        &
        \textbf{Architecture}
        &
        \textbf{LR}
        &
        \textbf{WD}
        &
        \textbf{Scheduler}
        &
        \textbf{Val RMSE}
        &
        \textbf{Fixed RMSE}
        &
        \textbf{Growth RMSE}
        \\
        \hline
        0
        &
        16-19-9
        &
        0.05
        &
        0.0
        &
        cosineannealing
        &
        0.0719
        &
        0.0705
        &
        0.0642
        \\
        1
        &
        9-18-18
        &
        0.04
        &
        0.001
        &
        cosineannealing
        &
        0.0660
        &
        0.0626
        &
        0.0680
        \\
        2
        &
        13-16-15
        &
        0.08
        &
        0.0
        &
        cosineannealing
        &
        0.0719
        &
        0.0749
        &
        0.0585
        \\
        3
        &
        11-17-17
        &
        0.08
        &
        0.001
        &
        cosineannealing
        &
        0.0687
        &
        0.0721
        &
        0.0544
        \\
        \hline
    \end{tabular}
\end{table*}

For completeness, the difference between the fixed retraining result and the
growth result in test RMSE is \(+0.0062\) for seed \(0\), \(-0.0055\) for
seed \(1\), \(+0.0164\) for seed \(2\), and \(+0.0176\) for seed \(3\).

These results show that the proposed train-and-grow algorithm can identify an
architecture together with a training trajectory whose test performance is
better on average than retraining the same structure from scratch with the
same seed. The difference does not imply a uniform advantage, since fixed
retraining wins for seed \(1\), but it shows that fixed retraining does not systematically improve the architectures
produced by CAGE-NAS. This behavior contrasts with the results commonly reported for other growth
algorithms, where retraining the final architecture after the growth process
usually produces a noticeable improvement in network performance.

\subsection{Construction of the synthetic \texorpdfstring{\(N=1024\)}{N=1024} benchmark}
\label{app:synthetic-n1024}

The synthetic benchmark is a deterministic smooth nonlinear regression task
with inputs
\[
x\in\mathbb{R}^4,
\qquad
x\sim\mathcal{N}(0,I_4).
\]

Only the first three coordinates contribute to the target. For the
single-output setting used in the experiments,
\[
y(x)
=
\frac{
\sin(1.5x_1)
+
\sin(2.25x_2)
+
\sin(3x_3)
}{3}
+
0.25\sin(x_1x_3)
+
\frac{0.1}{3}(x_1+x_2+x_3).
\]

Thus, the task combines smooth nonlinear terms, a pairwise interaction, and a
weak linear component, while \(x_4\) acts as an irrelevant input dimension.
No observation noise is added.

The training and validation sets contain \(1024\) examples each, generated as
\(16\) batches of \(64\) samples, while the test set contains \(8192\) examples.
The corresponding seeds are \(0\), \(2\), and \(1\). The synthetic loader
resets its seed at every iteration, so the same examples are observed at each
epoch.

\subsection{Fixed-architecture grid search}
\label{app:fixed-architecture-grid}

To determine whether the architectures discovered by CAGE-NAS could achieve
comparable performance under conventional parametric training, the four final
architectures
\[
[16,19,9],\quad
[9,18,18],\quad
[13,16,15],\quad
[11,17,17]
\]
were retrained from random initialization with structural growth disabled.

All runs use the same synthetic data split described in
Appendix~\ref{app:synthetic-n1024}. The grid evaluates \(60\) AdamW
configurations per architecture and seed over learning rate, weight decay, and
the presence or absence of cosine annealing. Four model seeds are evaluated
for each architecture over \(400\) epochs, resulting in \(960\) trials.

For the RMSE comparison, hyperparameters and the best epoch are selected using
minimum validation MSE (equivalently, minimum validation RMSE) only. The
primary comparison with CAGE-NAS uses paired same-seed retraining:
for each growth seed \(s\), the architecture discovered with that seed is
reinitialized using the same model seed \(s\), trained from scratch, and
compared with the corresponding growth run only after validation-based model
selection.

\subsection{Search over fixed architectures within the parameter budget}
\label{app:smaller-architecture-grid}

A second fixed-architecture experiment tests whether admissible architectures
within the corresponding parameter budget can obtain equal or better
performance under conventional training.

For a three-hidden-layer architecture
\[
A=[h_1,h_2,h_3],
\]
with four inputs and one output, the number of trainable parameters is
\[
p(A)
=
5h_1+h_1h_2+h_2+h_2h_3+2h_3+1.
\]

For each growth seed \(s\), architectures are enumerated independently under
the constraint
\[
400
\leq
p(A)
<
B_s,
\]
with hidden-layer widths constrained to be at least \(2\), where the corresponding
CAGE-NAS parameter budgets \(B_s\) are \(620\), \(624\), \(626\), and \(602\)
parameters for seeds \(0\) through \(3\), respectively.

The search follows a two-stage protocol. In Stage~1, every admissible
architecture is trained once from fresh weights using the same AdamW screening
configuration and the model seed corresponding to the growth run. The best
\(50\) architectures per seed according to validation performance are retained.
In Stage~2, these candidates are evaluated over \(60\) AdamW
learning-rate, weight-decay, and scheduler configurations.

For the reported RMSE comparison, candidates satisfying \(p(A)\leq600\) are
ranked by minimum validation MSE (equivalently, minimum validation RMSE), and
the corresponding test RMSE is read only after selection. Test performance is
not used during architecture or hyperparameter selection. Consequently, this
experiment tests whether an admissible architecture within the parameter
budget can outperform the architecture produced by growth within the evaluated
search protocol, rather than claiming the non-existence of a better
architecture outside the explored space.

\section{CAGE-NAS train-and-grow algorithms}
\label{app:train-and-grow-algorithms}

\begin{algorithm}[H]
\footnotesize
\setlength{\abovedisplayskip}{2pt}
\setlength{\belowdisplayskip}{2pt}
\setlength{\abovedisplayshortskip}{2pt}
\setlength{\belowdisplayshortskip}{2pt}
\setlength{\jot}{2pt}
\caption{Epoch-level certified adaptive training}
\label{alg:epoch-level-certified-training}
\begin{algorithmic}[1]

\Require Initial architecture \(A_0\), parameters \(\theta_0\),
training set \(\mathcal D_{\mathrm{tr}}\), validation set
\(\mathcal D_{\mathrm{val}}\), test set \(\mathcal D_{\mathrm{te}}\),
number of epochs \(E\), maximum outer steps
\(K_{\mathrm{out}}=15\)

\Ensure Final architecture and parameters \((A,\theta)\)

\State \((A,\theta)\gets(A_0,\theta_0)\)

\For{\(e=1,\ldots,E\)}

    \State Construct the fixed empirical certification set
    \[
        (X,Y)\gets
        \Call{BuildCertificationSet}{\mathcal D_{\mathrm{tr}}}
    \]

    \State \(n_{\mathrm{accepted}}\gets0\)
    \State \(s_{\mathrm{last}}\gets\textsc{None}\)
    \State \(\varepsilon_{\mathrm{last}}\gets+\infty\)

    \For{\(j=1,\ldots,K_{\mathrm{out}}\)}

        \State
        \[
        (A',\theta',s_j,\varepsilon_j)
        \gets
        \Call{CertifiedOuterStep}
        {A,\theta,X,Y,\mathcal D_{\mathrm{val}}}
        \]

        \Comment{Function-preserving growth performed inside the
        outer step is retained}

        \State \((A,\theta)\gets(A',\theta')\)
        \State \(s_{\mathrm{last}}\gets s_j\)
        \State \(\varepsilon_{\mathrm{last}}\gets\varepsilon_j\)

        \If{\(s_j=\textsc{AcceptedTangent}\)}
            \State
            \(n_{\mathrm{accepted}}
            \gets n_{\mathrm{accepted}}+1\)

        \ElsIf{\(s_j=\textsc{AcceptedNonlinear}\)}
            \State
            \(n_{\mathrm{accepted}}
            \gets n_{\mathrm{accepted}}+1\)

        \Else
            \State \textbf{break}
            \Comment{stop at the first rejected outer-step attempt}
        \EndIf

    \EndFor

    \State Evaluate epoch metrics on
    \(\mathcal D_{\mathrm{tr}}\),
    \(\mathcal D_{\mathrm{val}}\), and
    \(\mathcal D_{\mathrm{te}}\)

    \State
    \[
    q_{\mathrm{ext}}
    \gets
    \Call{ExternalGrowthDecision}
    {
        A,\theta,
        n_{\mathrm{accepted}},
        s_{\mathrm{last}},
        \varepsilon_{\mathrm{last}},
        \mathcal D_{\mathrm{tr}},
        \mathcal D_{\mathrm{val}}
    }
    \]

    \If{\(q_{\mathrm{ext}}=\textsc{True}\)}

        \State
        \[
        (A,\theta)
        \gets
        \Call{ExpressivityBottleneckGrowth}
        {A,\theta,\mathcal D_{\mathrm{tr}}}
        \]

    \EndIf

\EndFor

\State \Return \((A,\theta)\)

\end{algorithmic}
\end{algorithm}

\begin{algorithm}[H]
\footnotesize
\setlength{\abovedisplayskip}{2pt}
\setlength{\belowdisplayskip}{2pt}
\setlength{\abovedisplayshortskip}{2pt}
\setlength{\belowdisplayshortskip}{2pt}
\setlength{\jot}{2pt}
\caption{One certified function-space outer step}
\label{alg:certified-outer-step}
\begin{algorithmic}[1]

\Require Architecture \(A\), parameters \(\theta\), fixed certification set
\((X,Y)\), validation set \(\mathcal D_{\mathrm{val}}\)

\Ensure Updated pair \((A',\theta')\), status \(s\), and final
relative error \(\varepsilon\)

\State \(F\gets F_{A,\theta}(X)\)
\State \(r\gets2(F-Y)\)

\State
\[
(A_c,\theta_c,\varepsilon_c,s_c)
\gets
\Call{GrowUntilCertified}{A,\theta,X,Y}
\]

\If{\(s_c=\textsc{AcceptedNonlinear}\)}

    \State
    \Return
    \(
        A_c,\theta_c,
        \textsc{AcceptedNonlinear},
        \varepsilon_c
    \)

\EndIf

\If{\(s_c=\textsc{Blocked}\)}

    \State
    \Return
    \(
        A_c,\theta_c,
        \textsc{Rejected},
        \varepsilon_c
    \)

\EndIf

\Comment{The tangent family is now admissible:
\(\varepsilon_c<1/2\)}

\State
\[
(\widehat\theta,s_{\mathrm{tan}},\varepsilon_{\mathrm{tan}})
\gets
\Call{CertifiedTangentRealization}
{
    A_c,\theta_c,X,Y,\mathcal D_{\mathrm{val}}
}
\]

\If{\(s_{\mathrm{tan}}=\textsc{Accepted}\)}

    \State
    \Return
    \(
        A_c,\widehat\theta,
        \textsc{AcceptedTangent},
        \varepsilon_{\mathrm{tan}}
    \)

\EndIf

\Comment{Keep any function-preserving structural growth,
but reject the parameter step}

\State
\Return
\(
    A_c,\theta_c,
    \textsc{Rejected},
    \varepsilon_{\mathrm{tan}}
\)

\end{algorithmic}
\end{algorithm}

\end{document}